\documentclass[letterpaper, 10 pt, journal, twoside]{IEEEtran}
\usepackage{graphicx}
\usepackage{cite}
\usepackage{subfigure}
\usepackage{multirow}
\usepackage{stfloats}
\usepackage[ruled,vlined]{algorithm2e}

\begin{document}

\title{DeltaCharger: Charging Robot with Inverted Delta Mechanism and CNN-driven High Fidelity Tactile Perception for Precise 3D Positioning}

\author{Iaroslav Okunevich$^{1}$, Daria Trinitatova$^{1}$, Pavel Kopanev$^{1}$, and Dzmitry Tsetserukou$^{1}$%
\thanks{$^{1}$The authors are with the Intelligent Space Robotics Laboratory, Space CREI, Skolkovo Institute of Science and Technology, Moscow, Russian Federation.
        {\tt \{iaroslav.okunevich, daria.trinitatova, pavel.kopanev, d.tsetserukou\}@skoltech.ru}}%
}
%
%



\maketitle

\begin{abstract}
DeltaCharger is a novel charging robot with an Inverted Delta structure for 3D positioning of electrodes to achieve robust and safe transferring energy between two mobile robots. The embedded high-fidelity tactile sensors allow to estimate the angular, vertical and horizontal misalignments between electrodes on the charger mechanism and electrodes on the target robot using pressure data on the contact surfaces. This is crucial for preventing short circuit. In this paper, the mechanism of the developed prototype and evaluation study of different machine learning models for misalignment prediction are presented. The experimental results showed that the proposed system can measure the angle, vertical and horizontal values of misalignment from pressure data with an accuracy of 95.46\%, 98.2\%, and 86.9\%, respectively, using a Convolutional Neural Network (CNN). DeltaCharger can potentially bring a new level of charging systems and improve the prevalence of mobile autonomous robots.
\end{abstract}


%
\IEEEpeerreviewmaketitle

\section{Introduction}

\IEEEPARstart{N}{owadays}, the number of mobile robots is constantly increasing. They are applied in different areas such as logistics \cite{hangrawler2020, phollower2020}, stocktaking \cite{warevision2020}, assembling furniture \cite{Ikeabot}, education \cite{arvin2019mona}, and etc. In the face of COVID-19, a lot of companies and research groups have created autonomous mobile robots for disinfection \cite{Violet2020,ultrabot}. 

However, all mobile robots have a limited operation time \cite{batterySchedule2020}, which defines the range of mobile robots and depends on their energy capacity. This factor is not significant for indoor robots because increasing working space can be compensated by a larger number of docking stations. On the other hand, for outdoor robots that explore an unknown area (e.g., coal mine rescue robots \cite{CoalRescue}), the operation time problem is crucial.

There are many studies on increasing the working range of mobile robots. One of the most effective approaches is to apply renewable energy as a power source of the robot, such as solar energy \cite{SolarCS}. This makes it possible to have an unlimited working range of mobile robots. However, there are disadvantages, such as sensitivity to external conditions and low charging speed. The most common way to boost the range of a mobile robot is to increase the battery capacity. However, in this case the size and weight of a mobile robot are also increasing dramatically. An alternative approach could be taken from the aviation and space industries. In order to increase the range of a plane, additional fuel tanks or refueling from a tanker are used. Similarly, a new mobile robot, which would be simpler and cheaper than the original one, could recharge the main robot and increase its range. 

\begin{figure}
\centering
\includegraphics[width=0.98\linewidth]{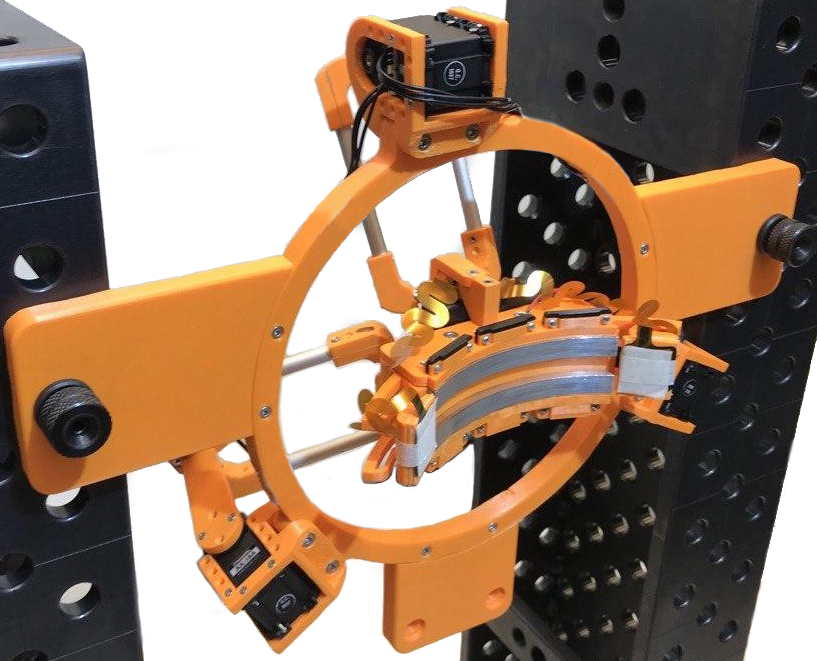}
\caption{The manufactured prototype of DeltaCharger.}
\label{DC}
\end{figure}
\subsection{Related works}

The idea of autonomous charging systems has been applied in some prototypes. The common approach is autonomous docking of the mobile robots with the charging station \cite{AutoDocking}. Another approach is the development of auto-charging robots \cite{CDAC,CDLC}. One of such charging systems is applied for the underwater robot \cite{fish}. The most advanced technology is using a mobile autonomous system for charging. The target object for the charging could be mobile devices \cite{phoneCharger1} or a swarm of robots \cite{swarmCharging}. Some concepts and prototypes of mobile robots for recharging larger systems exist, however, they are applied for working with electric vehicles in public places \cite{AMCPP}. The most famous are ``Laderoboter"\footnote{https://www.volkswagen-newsroom.com/en/stories/volkswagen-lets-its-charging-robots-loose-5700} by Volkswagen with a robotic arm (Universal Robot) as actuator and ``EVAR"\footnote{https://www.evar.co.kr/eng/pro04.php} by Samsung spin-off, both are fully autonomous.

Wireless \cite{soccer} or conductive types of charging implement the energy transmission in the charging system. Advantages of wireless power transmission include eliminating the electrical contact wear and providing isolation from the environment \cite{WirAdv}. However, the conductive charging system is more suitable for charging mobile robots, since powerful DC charging system allows to achieve higher power transmission performance than the wireless system \cite{AuCS}. 

The actuator of the charging system has to dock with the target robot. The size of the actuator defines the size of the whole autonomous charging robot. Thus, it has to be compact and should be placed inside the mobile platform. Applying a robotic arm as an actuator helps to dock with the main robot at a distance of about 0.2-1.2 meters. However, it will significantly increase the cost, size, weight, and power supply of the charging robot. Another solution is using a Cartesian manipulator as it is implemented in ``EVAR". However, it requires a lot of space and has a complicated structure for manufacturing. The Delta mechanism is more convenient to manufacture than the Cartesian mechanism. On the other hand, it requires more space inside the robot. An Inverted Delta actuator developed by us is more compact than the traditional Delta actuator and can be easily integrated inside the robot body. 

The main requirement for charging is a reliable and safe connection between electrodes. The angular misalignment between the electrodes on the charger and the target robot leads to a short circuit. The horizontal or vertical misalignment leads to a decrease in the contact area of the electrodes, which can cause damage to them. The same problem is solved by misalignment-sensing coils in inductive power transfer systems \cite{AutAlign}. The tactile perception system can potentially solve the misalignment issue in the conductive case.

In the current work, we propose DeltaCharger, an Inverted Delta actuator for autonomous charging of the mobile robot, which can provide a robust connection between the electrodes through the use of high-fidelity tactile sensors. The prototype of the DeltaCharger is shown in Fig. \ref{DC}.

\section{Design of DeltaCharger}
\subsection{Overall structure}
\begin{figure}[ht]
    \centering
    \includegraphics[width=0.97\linewidth]{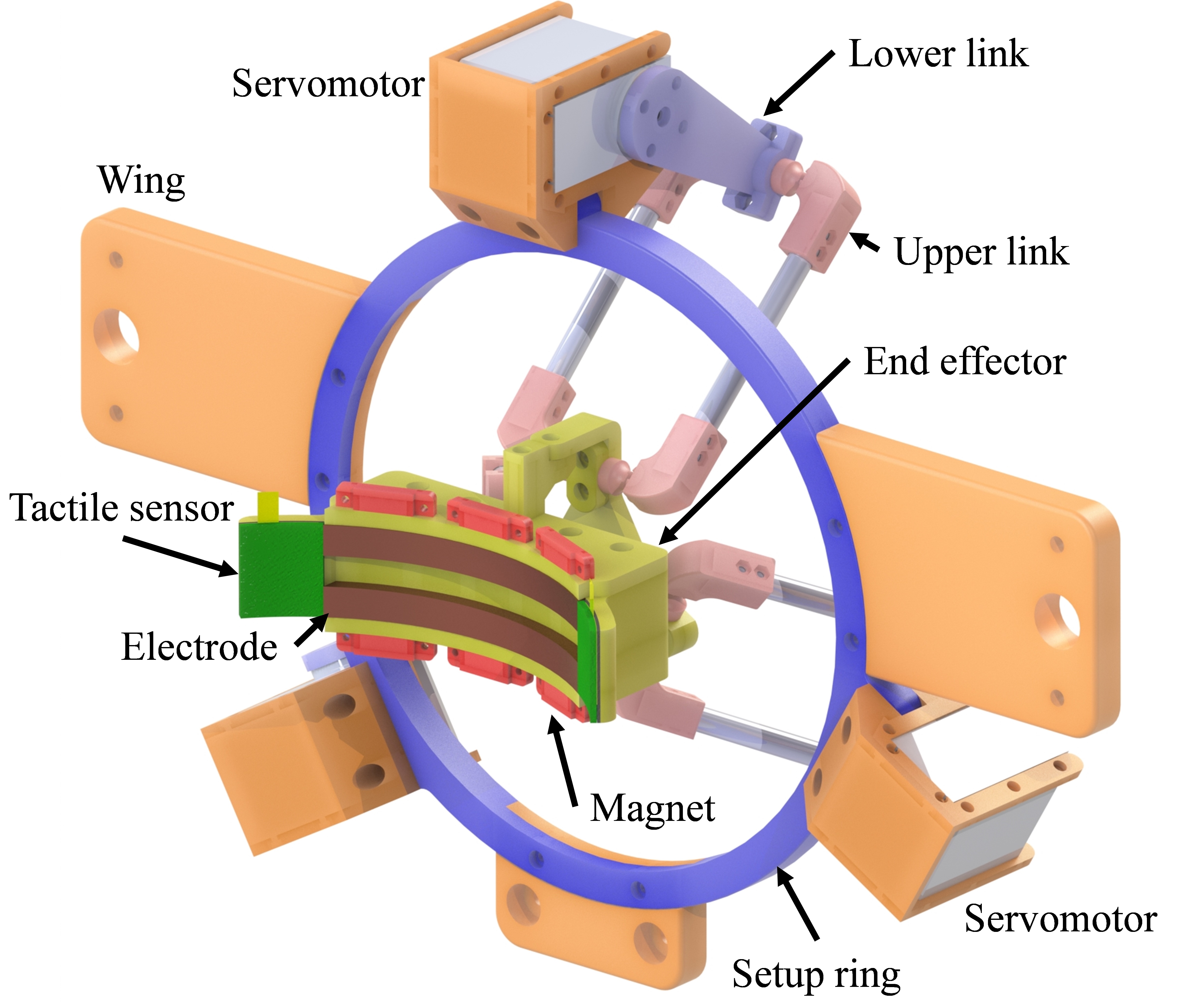}
    \caption{A CAD model of DeltaCharger.}
    \label{RDC}
\end{figure}

The design of DeltaCharger was inspired by DeltaTouch haptic interface \cite{touchvr}. The DeltaCharger mechanism consists of a setup ring with motors and a moving platform with two electrodes (end effector). The setup ring is mounted on a mobile charging robot using three wings. The mechanism is actuated by three servomotors (Dynamixel MX64). The Inverted Delta mechanism comprises three interchangeable kinematic limbs that allow the moving of the end effector as shown in Fig. \ref{RDC}. The actuator is manufactured from polylactic acid (PLA) material and hollow aluminum tubes.

The end effector consists of a platform with two electrodes, six magnets, and two tactile sensors. Electrodes are made of duralumin for greater strength. They withstand a current of 18 $A$.  Magnets keep the charger's electrodes in constant contact with a discharged robot. Tactile sensors are applied to detect a misalignment between the electrodes on the end effector and electrodes on the main robot. The weight of the end effector is of 274 $g$, and the weight of the whole actuator with the end effector equals 938 $g$. The working range of the end effector is of 60 $mm$ along the $X$, $Y$ axes in positive and negative directions and 110 $mm$ along the $Z$-axis in the positive direction.

\subsection{System configuration}

\begin{figure}[ht]
\centering
\includegraphics[width=0.95\linewidth]{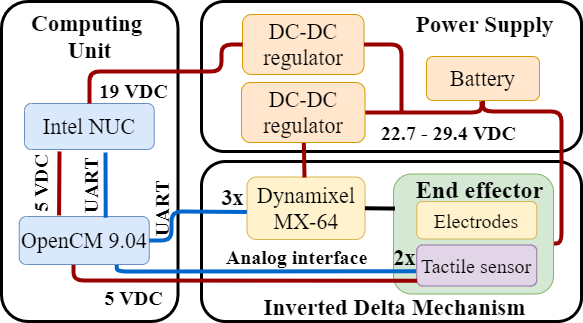}
\caption{The system configuration of DeltaCharger.}
\label{UBFD}
\end{figure}

The system configuration of DeltaCharger with 3 blocks is shown in Fig. \ref{UBFD}. The first block, i.e., Computing Unit, includes Intel NUC computer and OpenCM 9.04 Dynamixel Controller. The communication between the Intel NUC and the microcontroller was implemented through UART. A computer solves inverse kinematics problem for Inverted Delta robot and analyzes the data from sensors. The second block is the Inverted Delta Mechanism. It is an actuator, which has a setup ring, servomotors, and end effector. OpenCM board controls servomotors through UART. Additionally, the controller collects data from tactile sensors by means of an analog interface. The power supply voltage for tactile sensors equals 5 VDC. Power Supply block consists of a battery and two DC-DC regulators. The output voltage of the battery equals 22.7-29.8 VDC. Servomotors require 12 VDC and Intel NUC requires 19 VDC. DC-DC regulators are applied for reaching these voltages. Electrodes in the end effector are electrified by the battery to transfer energy. 

\subsection{Embedded sensors}
\begin{figure}[!ht]
\begin{center}
\subfigure[\label{misaligZero} Case of aligned electrodes.]{
\includegraphics[width=0.88\linewidth]{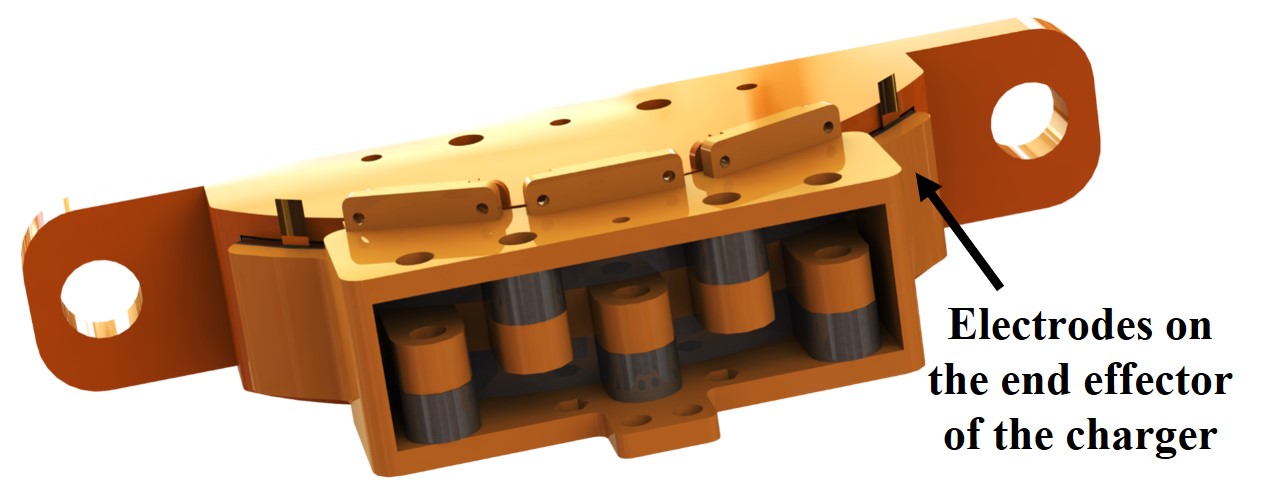}}
\subfigure[\label{misaligfour} Case of misaligned electrodes.]{
\includegraphics[width=0.81\linewidth]{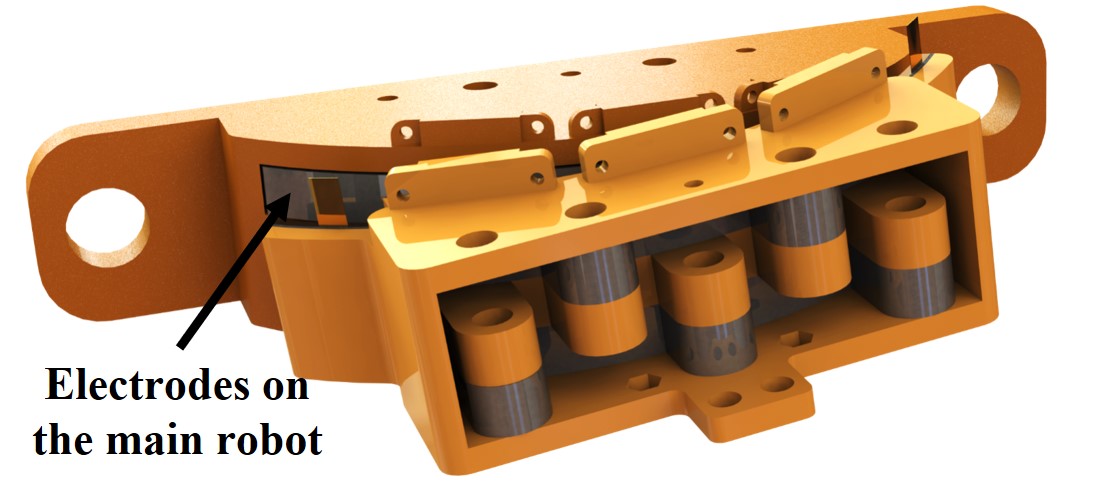}}
\caption{Docking between the electrodes of the main robot and DeltaCharger.}
\label{tilted angle}
\end{center}
\end{figure}

Mobile robots often perform their tasks on uneven surfaces. This can cause problems during the charging process because electrodes on the end effector of the charger must have the same orientation as electrodes on the target robot. If the angle between these electrodes is too large, the process of charging will not be effective. In a normal case, each electrode of DeltaCharger contacts with one electrode on the main robot, whereas in a tilt case, each electrode can contact with two electrodes, leading to a short circuit (Fig. \ref{tilted angle}). To solve this problem, one can measure the angle of misalignment between electrodes. If the angle is lower than a critical angle, charging can take place.

\begin{figure}[!t]
    \centering
    \includegraphics[width=0.9\linewidth]{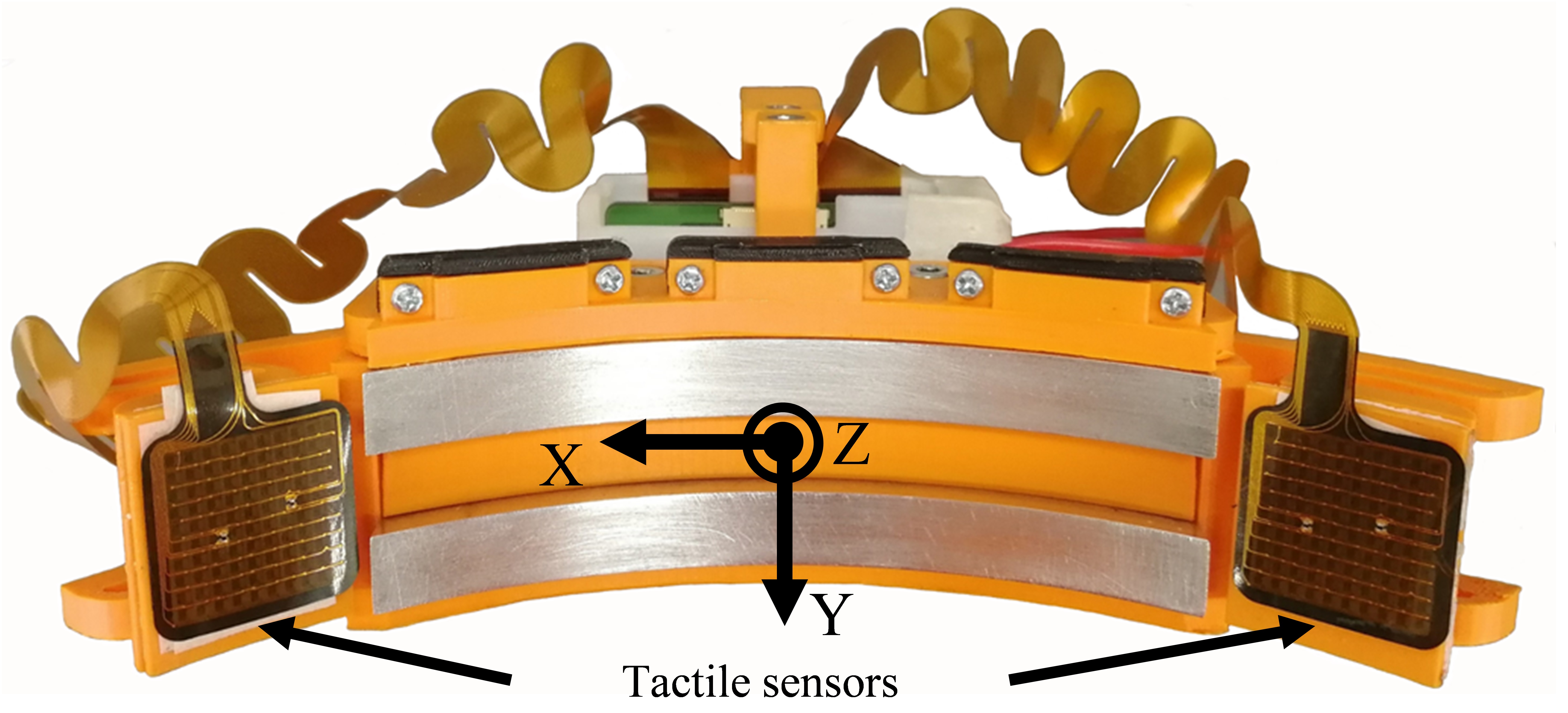}
    \caption{End effector with two embedded tactile sensor arrays.}
    \label{Sensor}
\end{figure}

To measure the angular misalignment, the system was equipped with tactile sensor arrays \cite{yem2019}. The end effector with tactile sensors is shown in Fig. \ref{Sensor}. Misalignment angle is the rotational angle around the $Z$-axis in the coordinate system of the actuator. Each array is capable of sensing the maximum frame area of $5.8\ cm^2$ with a resolution of 100 points per frame. The sensing frequency equals $120\ Hz$ (frames per second). With these sensors, the system can precisely detect the force applied to the edges of the charger's terminal. The range of force detection is of 1-9 $N$. The high resolution of the sensors allows to make a correlation between the obtained force diagrams and different misalignment angles. As a result, the prediction system can be developed to detect the angles of misalignment by contact patterns.

In addition, the docking precision of end effector is not high. The coordinates of the required position for the end effector are received from the computer vision (CV) system. It includes an RGB-D camera and detects the electrodes on the main robot. After that, it returns the coordinates of the electrodes in the actuator coordinate system. The disadvantage of this method is the working range for depth camera. The CV system cannot detect the coordinates of the docking point with high precision in proximity to the target robot. In this case, the electrodes may not touch, or the contact will not be over the entire area, leading to rapid wear of the electrodes. 
We propose to navigate the end effector after docking by a tactile system that is used to predict the misalignment angle. The horizontal and vertical misalignments would have different contact patterns, which depend on the value of misalignment. Machine Learning (ML) methods could be applied to predict the value of the end effector shifting to correct its position. 

\section{Tactile Perception of the End Effector Misalignment}

\subsection{Strategy of operation}
DeltaCharger will be part of a mobile charging robot with the following operating strategy. The mobile charger will wait for the message with coordinates of the discharged robot from the robot itself or from the base station. After that, it will move to the area of obtained coordinates and will start to search for the electrodes by the RGB-D camera-based CV system. If the charging robot cannot find the electrodes, it should return to the base station. If the electrodes are recognized, the charger moves to the electrodes and docks with them according to Algorithm \ref{safe_alg}.

\begin{algorithm} [!h] 
\caption{\emph{Algorithm of safe docking with the target robot.}}
\label{safe_alg}
receive the coordinates of electrodes from the visual system\;
move the end effector to the received position\;
\If{the consuming current of servomotors is more than 0.5 $A$}{

    decrease the $Z$-coordinate of the end effector on 3 $cm$
    }
\While{current consumption of servomotors is less than 0.4 $A$}{

    increment the $Z$-coordinate of the required position on 5 $mm$
    }
measure the misalignment angle\;
\eIf{the misalignment angle is less than the critical angle}{
  \While{the horizontal and vertical misalignments are not equal to zero}{
  
        detect the value of misalignments\;
  
        move the end effector to align it\;
         
        }
    start charging\;
    }
    {
            return the end effector to the initial position\;
            
            charging is failed\;
    }
\end{algorithm}

The adjustment of the end effector position along the $Z$-axis is carried out by measuring the value of the current consumed from the motors. The position of the electrodes calculated by the CV system can be farther than the real position of the electrodes. In this case, the end effector leans on the electrodes with a high torque value, and the consuming current reaches a value of several $A$. The current over 0.5 $A$ is too high for the continuous work of the servomotors, which will lead to overheating after a couple of minutes of operation. The prediction of the end effector position error by consuming current is not sensitive enough. A reliable method is to move the end effector back a few $cm$ and move it forward with small steps. If the consuming current is higher than 0.4 $A$ and lower than 0.5 $A$, the end effector rests on the electrodes, and servomotors can hold the end effector for 4 hours without overheating. The tactile sensors are not applied to detect the error of end effector position along the $Z$-axis since the difference between the obtained pressure data for high and normal consuming current is not large. Thus, it is not possible to predict a case when the servomotors are close to overheating.

After the end effector is aligned along $Z$-axis, the misalignment angle is measured by the tactile system. The end effector will be equipped with a rotational mechanism to align the end effector with the electrodes in future work. In the current work, if the misalignment angle is more than the critical one, the charging process will be stopped.

The last step of the algorithm is the end effector alignment along the $X$ and $Y$ axes. The tactile system detects the values of vertical and horizontal misalignments. After that, the end effector moves to decrease the misalignment value. When the misalignment values equal zero or the alignment operation has been repeated more than 15 times, the end effector alignment process is over, and the charging process is started. The limitation of 15 loop executions is needed to prevent an infinite alignment cycle.

\subsection{End effector misalignment prediction}
In order to estimate the angular, vertical, and horizontal misalignments, a classification problem was solved. Our first step was to find the roll angle constraints and determine the possible classes (misalignment angle ranges). Outdoor robots can tilt more than 30$^{\circ}$ because they have to cross the rough terrain, while for indoor robots angle constraints are different. To determine the maximum possible roll angle of the robot, we have conducted the experiment with an indoor robot.

In our case, the robot had a maximum tilt angle of 5$^{\circ}$. The critical angle for the shape of electrodes is 12$^{\circ}$. It is not enough for charging very tilted outdoor robots. However, it is sufficient for a target indoor robot. Therefore, 6 classes of angular misalignment were defined with a range of 1$^{\circ}$. Furthermore, predicting vertical and horizontal misalignments has been identified as a classification problem. Ideally, the actuator can rotate the end effector so that the misalignment angle would be 0$^{\circ}$ after the successful detection. In this case, the prediction of the position misalignment had to be calculated only for the angle of 0$^{\circ}$. The pressure input data from tactile sensors do not have such strong differences to detect the misalignment with a precision of 1 $mm$. However, the misalignment with the step of 5 $mm$ can be detected by contact patterns. The error of the electrode detection by the camera in proximity to the target robot equals 1 $cm$. Thus, we were changing the end effector position to obtain contact patterns for horizontal and vertical misalignments from -10 to 10 $mm$ with a step of 5 $mm$. Therefore, 5 classes of misalignment for each direction were defined. If absolute vertical or horizontal misalignment is more than 10 $mm$, tactile sensors cannot touch the electrodes of the main robot after the docking. In the case of receiving zero input arrays from tactile sensors, the charging process will end.

For solving the classification problem and determining the optimal model for the misalignment prediction, shallow and Deep Learning (DL) approaches were tested. 

\subsection{Shallow learning approach}

DeltaCharger has two tactile sensor arrays with a resolution of 100 points per frame each. Therefore, the dimension of the input vector for the classification problem is 200. We have explored 7 popular classification models:

\begin{enumerate}
	\item Linear classification models:
    \begin{itemize}
    	\item Logistic Regression (LR) Classifier with cross-validation estimator;
        \item Support Vector Machine (SVM) with stochastic gradient descent (SGD) learning.
    \end{itemize}
    \item A supervised neighbors-based learning methods:
    \begin{itemize}
    	\item K-nearest neighbors (KNN) Classifier.
    \end{itemize}
    \item Decision tree non-parametric supervised models:
    \begin{itemize}
    	\item Decision Tree (DT) Classifier.
    \end{itemize}
    \item Ensemble methods:
    \begin{itemize}
    	\item Gradient Boosted Decision Tree (GBDT) Classifier;
    	\item Random Forest (RF) Classifier.
    \end{itemize}
    \item Extreme Gradient Boosting model:
    \begin{itemize}
    	\item XGBoost Classifier \cite{xgboost}.
    \end{itemize}
\end{enumerate}

\subsection{Deep learning approach}

\begin{figure}[!ht]
\begin{center}
\subfigure[\label{RNN}Regular NN architecture.]{
\includegraphics[width=0.92\linewidth]{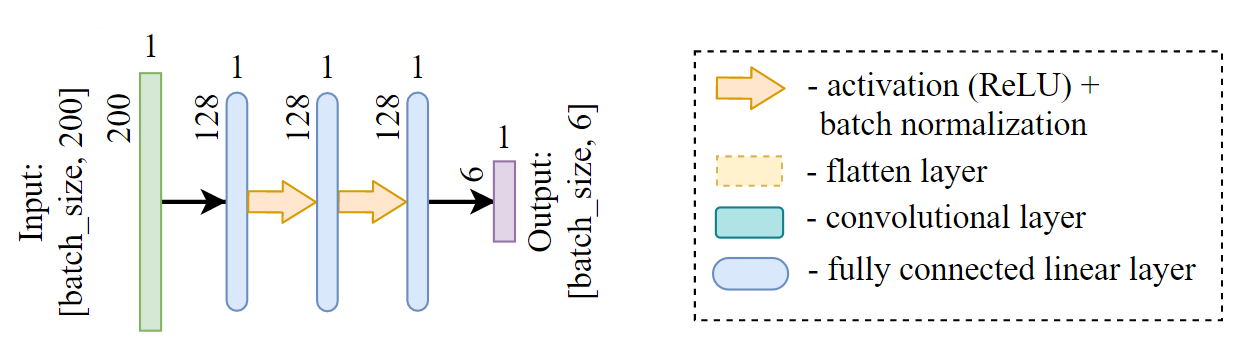}}
\subfigure[\label{CNN}CNN architecture.]{
\includegraphics[width=0.92\linewidth]{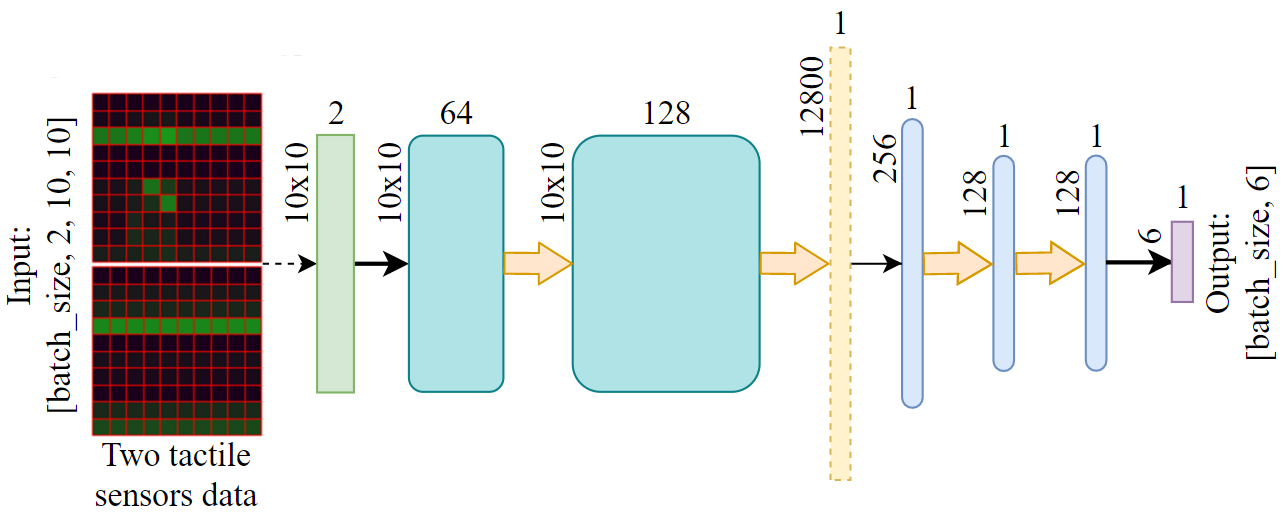}}
\caption{Developed deep learning architectures.}
\label{ARCH}
\end{center}
\end{figure}

Deep Learning (DL) methods apply non-linear processing layers for discriminative or generative feature abstraction and transformation for pattern analysis. We can use a regular neural network (NN) with an input vector of 1x200 or transform the tactile data to a 10x10 array with 2 channels. This transformation allows to represent the obtained force values as an image and apply a convolutional neural network (CNN) for the prediction. We developed the regular NN and CNN to compare them with each other and to achieve a better performance than shallow learning methods. The developed architectures are shown in Fig. \ref{ARCH}.

The regular NN consists of fully connected linear layers with ReLU (Rectified Linear Unit) nonlinear activation functions and batch normalization. Batch normalization acts as a regularizer and speeds up the training of classification models \cite{batch}. In addition, with batch normalization, gradients used in training are more predictive and well-behaved, which enables faster and more effective optimization \cite{batch_why}. CNN includes 2 convolutional layers with a 3x3 kernel and 3 fully connected linear layers with ReLU nonlinear activation functions and batch normalization.

\section{Experiments}
\subsection{Sensor data}

\begin{figure}[ht!]
    \centering
    \includegraphics[width=0.48\textwidth]{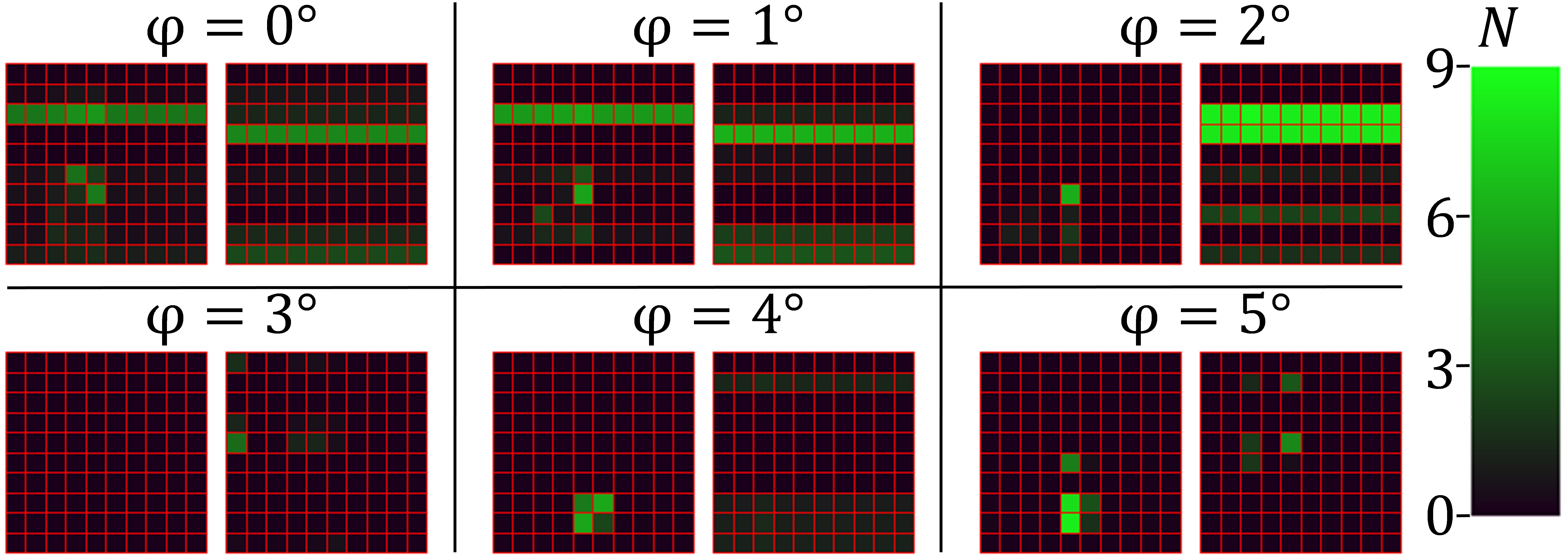}
    \caption{Tactile patterns for different misalignment angles $\varphi$. The color saturation of a cell corresponds to the force applied to it: black and bright green cells represent 0 $N$ and  9 $N$, respectively.}
    \label{PAT}
\end{figure}

For training and validation of classification models, 2 datasets were collected. The first one includes 600 data pairs from tactile sensor arrays. The main robot was tilted at an angle from 0 to 5$^{\circ}$. Per each angle span of 1$^{\circ}$, the end effector was docked with the main robot, and measurements were collected from 2 tactile sensors 100 times. Therefore, the collected data was balanced. The initial position of the end effector was varied from -20 to 20 $mm$ along the $X$ and $Y$ axes with a step of 4 $mm$. Fig. \ref{PAT} shows a set of obtained patterns for different values of misalignment angle $\varphi$. 

The second dataset includes 500 data pairs from tactile sensor arrays. The actuator was installed opposite to the counterpart with electrodes so that the coordinates of the electrodes in the actuator coordinate system were equal to (0; 0; 80) (coordinates are given in millimeters) at a zero misalignment angle. After that, the robot performed 20 attempts to dock with the electrodes. In a similar way, the target point for the actuator was changing with a 5 $mm$ increment along the $X$ and $Y$ axes. As a result, the coordinates of the end effector changed from (-10; -10; 80) to (10; 10; 80).

Both datasets were split into training (67\% of data) and validation sets (33\% of data) to avoid resubstitution validation or resubstitution evaluation \cite{raschka2018model}. For shallow learning models and regular NN, the data was flattened to one-dimensional vectors with 200 elements. For CNN, input data consisted of tensors with 2 channels (since we had two data pairs per sensor measurement) and with the shape of 10x10.

\subsection{Comparison of classification models for misalignment prediction}
All classification models were trained on the Intel NUC Mini PC (parameters are listed in the Table \ref{tabl:T1}).

\begin{table}[h]
    \centering
    \caption{Parameters of the Computer}
    \label{tabl:T1}
    \scalebox{0.88}{%
    \begin{tabular}{|c|c|}
    \hline
     Computing unit & Intel NUC NUC8i7BEH \\
    \hline
    \multirow{2}{*}{Processor}  &  Intel Core(TM) i7-8559U \\ 
     & (2.7 GHz – 4.5 GHz, Quad Core, 8MB Cache, 28W TDP) \\
    \hline
     Memory & Two DDR4 SO-DIMM sockets (32 GB, 2400 MHz)   \\
    \hline
     Operating system &  Ubuntu 18.04.5 LTS \\
    \hline
    \end{tabular}}
\end{table}

\subsubsection{Classification models}

All shallow learning models except XGBoost, which has its own python package, were trained using the Scikit-learn\footnote{https://scikit-learn.org/} machine learning library.

DL models were constructed using a PyTorch\footnote{https://pytorch.org/} open-source machine learning framework. The cross-entropy loss was used as the criterion. It combines the LogSoftmax function with the negative log-likelihood loss. This criterion is commonly used for multilabel classification problems. Besides, we used the dynamic learning rate reducing scheduler ReduceLROnPlateau. Models often benefit from reducing the learning rate by a factor of 2-10 once learning stagnates. DL models for each case were trained during 50 epochs.

\subsubsection{Combined results for angular misalignment prediction}

\begin{figure}[!ht]
    \centering
    \includegraphics[width=0.92\linewidth]{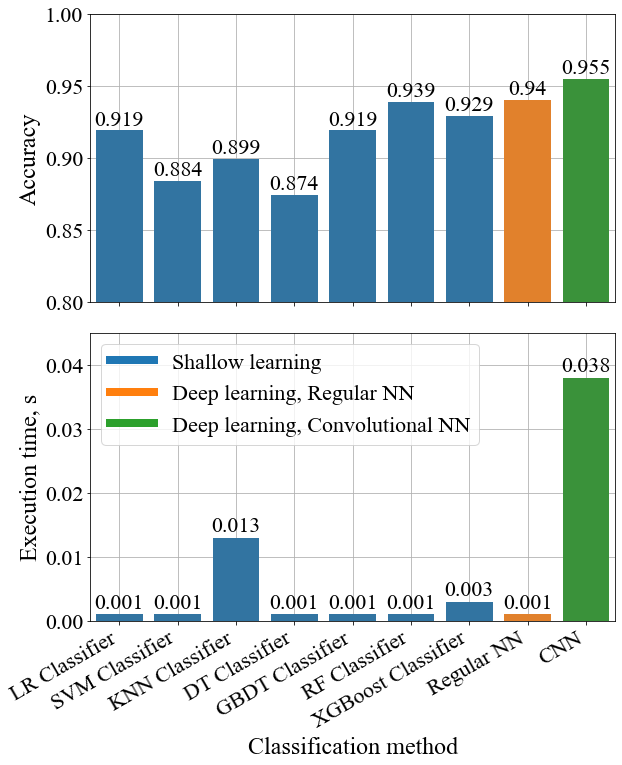}
    \caption{Accuracy and execution time comparison of all explored classification models for the angle prediction.}
    \label{Accuracy}
\end{figure}

The experimental results of all applied classification models for the misalignment angle prediction are shown in Fig. \ref{Accuracy}. 

The collected datasets from two tactile sensors were balanced. Therefore, we chose accuracy as a quality metric for the final evaluation of all models for misalignment prediction. RF Classifier showed the best performance for the end effector misalignment detection between all shallow learning models. DL methods showed higher accuracy than shallow learning approaches, and CNN had the highest accuracy of 95.46\% in the misalignment angle prediction problem. The disadvantage of CNN is the lowest speed among all models. Considering the speed of each method, the CNN model was significantly slower than the other models with an execution time of 38 $ms$. However, this time is still acceptable for the docking task.

In conclusion, with CPU utilization, RF Classifier and regular NN showed similar performance considering model accuracy and execution speed. However, taking into account model learning time, which was $0.017\ s$ and $6.597\ s$ for RF Classifier and regular NN, respectively, in the misalignment angle prediction problem with CPU utilization, RF Classifier showed the best overall performance among all classification methods. However, in the next iteration of this project, we are going to use NVIDIA Jetson Nano with GPU acceleration. Therefore, CNN is an optimal classification model in this regard. It provides higher accuracy than other models, and with GPU acceleration it will be significantly faster. 

\subsubsection{Combined results for vertical and horizontal misalignment prediction}

\begin{figure}[!h]
\begin{center}
\subfigure[\label{vertical_result}Accuracy of models for the vertical misalignment prediction.]{
\includegraphics[width=0.93\linewidth]{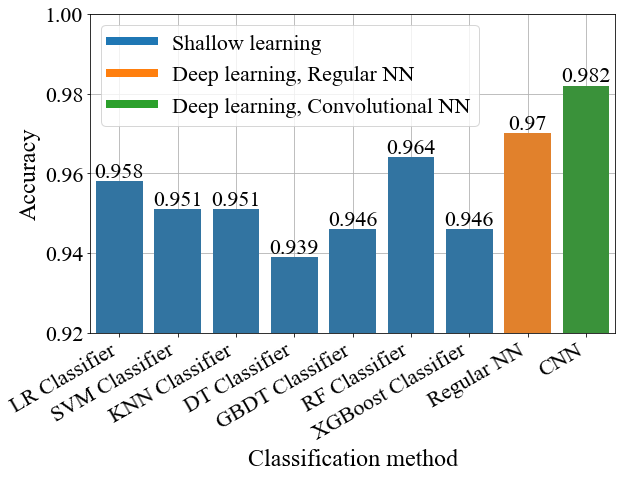}}
\subfigure[\label{horizontal_result}Accuracy comparison for the horizontal misalignment prediction.]{
\includegraphics[width=0.93\linewidth]{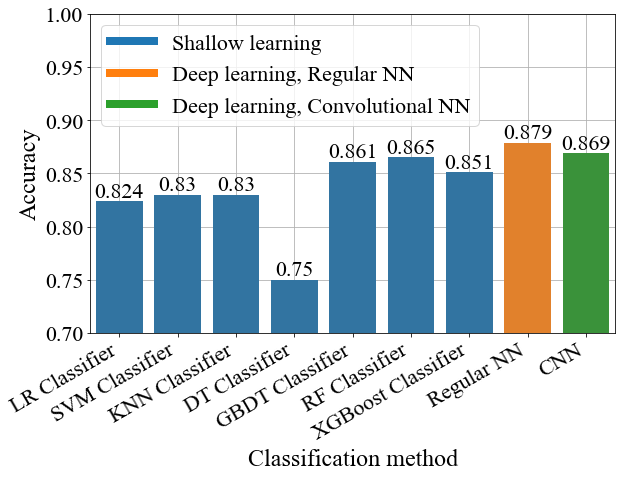}}
\caption{Accuracy comparison of all explored classification models for the misalignment prediction.}
\label{2exp_result}
\end{center}
\end{figure}

The DL and shallow learning models were retrained on the second dataset with the values of horizontal and vertical misalignment. The experimental results of all applied classification models for vertical and horizontal misalignment prediction are shown in Fig. \ref{2exp_result}.

The DL models showed higher accuracy than shallow learning approaches. However, the performance of the former for the prediction of vertical and horizontal misalignments is different. CNN showed a similar performance as ensemble methods, and regular NN had the highest accuracy of 87.9\% in the horizontal misalignment prediction. For the prediction of vertical misalignment, CNN showed the highest accuracy of 98.2\%.

\subsection{Discussion}

The CNN-based angular and vertical misalignment detection had the maximum accuracy of 95.46\% and 98.2\%, respectively. And for the detection of the horizontal misalignment, regular NN showed the highest accuracy of 87.9\%. The force patterns mainly represented the positions of the electrodes of the main robot, which were pressing on the area of the sensor. The accuracy of predicting the angular and vertical misalignments was high due to differences in contact patterns depending on the position of the electrodes. The distribution of pressure points on the electrodes changed with the end effector orientation and shifted up or down on the obtained patterns.
In the case of horizontal misalignment, the main difference between patterns was the force magnitude, while the shape of the patterns did not differ much. It is the main reason why detection of horizontal misalignment had a lower accuracy. One approach to solve this problem is applying the special shape of the electrode on the main robot, which can help to create unique patterns for horizontal misalignment. For example, if the electrodes have holes visible on the contact patterns, then the horizontal displacement will be easier to predict.

\section{Conclusions and Future Work}

In this work, we presented DeltaCharger, the Inverted Delta manipulator with embedded high-fidelity tactile sensors for charging mobile robots. The Inverted Delta mechanism is proposed to achieve a compact and reliable structure. The CNN-driven tactile perception system using data from tactile pressure sensors can accurately recognize the angular, vertical, and horizontal misalignments between the electrodes of the charger and the electrodes of the main robot. Based on the estimated angle, the proposed system decides whether it is safe to charge the main robot or the current position of the electrodes will cause a short circuit. Depending on the vertical and horizontal misalignments, the system adjusts the end effector position to reach the entire area of contact between the electrodes. The proposed CNN-based tactile perception system can predict the angle, vertical, and horizontal values of end effector misalignment from pressure data with an accuracy of $95.46\%$, $98.2\%$, and $86.9\%$, respectively. DeltaCharger can also be applied in charging of the heterogeneous robots, such as, mobile, quadruped, UAV, with different height of charging electrodes.

The future work will be aimed at developing and assembling a mobile robot for charging. Besides, the mechanism of the end effector will be improved to allow electrode rotating in order to make it possible to charge the mobile robot tilted at the critical angle. 

\section*{Acknowledgment}

The authors would like to thank Professor Hiroyuki Kajimoto, University of Electro-Communications, Japan for providing tactile sensor arrays, and researcher Jonathan Tirado, Skolkovo Institute of Science and Technology, for helping with setting up the tactile sensors.

\bibliographystyle{IEEEtran} 
\bibliography{references}

\begin{thebibliography}{10}
\providecommand{\url}[1]{#1}
\csname url@samestyle\endcsname
\providecommand{\newblock}{\relax}
\providecommand{\bibinfo}[2]{#2}
\providecommand{\BIBentrySTDinterwordspacing}{\spaceskip=0pt\relax}
\providecommand{\BIBentryALTinterwordstretchfactor}{4}
\providecommand{\BIBentryALTinterwordspacing}{\spaceskip=\fontdimen2\font plus
\BIBentryALTinterwordstretchfactor\fontdimen3\font minus
  \fontdimen4\font\relax}
\providecommand{\BIBforeignlanguage}[2]{{%
\expandafter\ifx\csname l@#1\endcsname\relax
\typeout{** WARNING: IEEEtran.bst: No hyphenation pattern has been}%
\typeout{** loaded for the language `#1'. Using the pattern for}%
\typeout{** the default language instead.}%
\else
\language=\csname l@#1\endcsname
\fi
#2}}
\providecommand{\BIBdecl}{\relax}
\BIBdecl

\bibitem{hangrawler2020}
R.~Fukui, Y.~Yamada, K.~Mitsudome, K.~Sano, and S.~Warisawa, ``Han{G}rawler:
  {L}arge-{P}ayload and {H}igh-{S}peed {C}eiling {M}obile {R}obot {U}sing
  {C}rawler,'' \emph{IEEE Transactions on Robotics}, vol.~36, no.~4, pp.
  1053--1066, 2020.

\bibitem{phollower2020}
J.~Ba{\v{c}}{\'\i}k, P.~Tk{\'a}{\v{c}}, L.~Hric, S.~Alexovi{\v{c}}, K.~Kyslan,
  R.~Olexa, and D.~Perdukov{\'a}, ``Phollower—{T}he {U}niversal {A}utonomous
  {M}obile {R}obot for {I}ndustry and {C}ivil {E}nvironments with {COVID}-19
  {G}ermicide {A}ddon {M}eeting {S}afety {R}equirements,'' \emph{Applied
  Sciences}, vol.~10, no.~21, p. 7682, 2020.

\bibitem{warevision2020}
I.~Kalinov, A.~Petrovsky, V.~Ilin, E.~Pristanskiy, M.~Kurenkov, V.~Ramzhaev,
  I.~Idrisov, and D.~Tsetserukou, ``Ware{V}ision: {CNN} {B}arcode
  {D}etection-{B}ased {UAV} {T}rajectory {O}ptimization for {A}utonomous
  {W}arehouse {S}tocktaking,'' \emph{IEEE Robotics and Automation Letters},
  vol.~5, no.~4, pp. 6647--6653, 2020.

\bibitem{Ikeabot}
R.~A. Knepper, T.~Layton, J.~Romanishin, and D.~Rus, ``Ikea{B}ot: {A}n
  autonomous multi-robot coordinated furniture assembly system,'' in \emph{2013
  IEEE International conference on robotics and automation}.\hskip 1em plus
  0.5em minus 0.4em\relax IEEE, 2013, pp. 855--862.

\bibitem{arvin2019mona}
F.~Arvin, J.~Espinosa, B.~Bird, A.~West, S.~Watson, and B.~Lennox, ``Mona: an
  affordable open-source mobile robot for education and research,''
  \emph{Journal of Intelligent \& Robotic Systems}, vol.~94, no.~3, pp.
  761--775, 2019.

\bibitem{Violet2020}
C.~McGinn, R.~Scott, N.~Donnelly, K.~L. Roberts, M.~Bogue, C.~Kiernan, and
  M.~Beckett, ``Exploring the {A}pplicability of {R}obot-{A}ssisted {UV}
  {D}isinfection in {R}adiology,'' \emph{Frontiers in Robotics and AI}, vol.~7,
  2020.

\bibitem{ultrabot}
S.~Perminov, N.~Mikhailovskiy, A.~Sedunin, I.~Okunevich, I.~Kalinov,
  M.~Kurenkov, and D.~Tsetserukou, ``Ultrabot: {A}utonomous {M}obile {R}obot
  for {I}ndoor {UV-C} {D}isinfection,'' in \emph{2021 17th International
  Conference on Automation Science and Engineering (CASE)}, 2021, in print.

\bibitem{batterySchedule2020}
M.~Tomy, B.~Lacerda, N.~Hawes, and J.~Wyatt, ``Battery charge scheduling in
  long-life autonomous mobile robots via multi-objective decision making under
  uncertainty,'' \emph{Robotics and Autonomous Systems}, 2020.

\bibitem{CoalRescue}
M.~Li, H.~Zhu, S.~You, L.~Wang, and C.~Tang, ``Efficient laser-based 3d slam
  for coal mine rescue robots,'' \emph{IEEE Access}, vol.~7, pp.
  14\,124--14\,138, 2018.

\bibitem{SolarCS}
K.~Nathaniel, L.~Yen-Chen, M.~Matthew, S.~Benjamin, B.~YunQi, and D.~Ran,
  ``Mission planning for a multi-robot team with a solar-powered charging
  station,'' \emph{2017 IEEE/RSJ International Conference on Intelligent Robots
  and Systems (IROS)}, pp. 5233--5238, 2017.

\bibitem{AutoDocking}
F.~Guangrui and W.~Geng, ``Vision-based autonomous docking and re-charging
  system for mobile robot in warehouse environment,'' in \emph{2017 2nd
  International Conference on Robotics and Automation Engineering
  (ICRAE)}.\hskip 1em plus 0.5em minus 0.4em\relax IEEE, 2017, pp. 79--83.

\bibitem{CDAC}
Y.~Lou and S.~Di, ``Design of a cable-driven auto-charging robot for electric
  vehicles,'' \emph{IEEE Access}, vol.~8, pp. 15\,640--15\,655, 2020.

\bibitem{CDLC}
H.~Yuan, Q.~Wu, and L.~Zhou, ``Concept design and load capacity analysis of a
  novel serial-parallel robot for the automatic charging of electric
  vehicles,'' \emph{Electronics}, vol.~9, no.~6, p. 956, 2020.

\bibitem{fish}
P.~Paul, J.~Cheong, and M.~Porfiri, ``An autonomous charging system for a
  robotic fish,'' \emph{IEEE/ASME Transactions on Mechatronics}, vol.~21,
  no.~6, pp. 2953--2963, 2016.

\bibitem{phoneCharger1}
W.~Yi-Shiun, C.-W. Chen, and H.~Samani, ``Development of wireless charging
  robot for indoor environment based on probabilistic roadmap,'' in
  \emph{International Conference on Interactive Collaborative Robotics}.\hskip
  1em plus 0.5em minus 0.4em\relax Springer, Cham, 2016, pp. 55--62.

\bibitem{swarmCharging}
F.~Arvin, K.~Samsudin, and A.~R. Ramli, ``Swarm robots long term autonomy using
  moveable charger,'' in \emph{2009 International Conference on Future Computer
  and Communication}.\hskip 1em plus 0.5em minus 0.4em\relax IEEE, 2009, pp.
  127--130.

\bibitem{AMCPP}
P.-Y. Kong, ``Autonomous robot-like mobile chargers for electric vehicles at
  public parking facilities,'' \emph{IEEE Transactions on Smart Grid}, vol.~10,
  no.~6, pp. 5952--5963, 2019.

\bibitem{soccer}
S.~C. Mukhopadhyay, G.~S. Gupta, and B.~J. Lake, ``Design of a contactless
  battery charger for micro-robots,'' in \emph{2008 IEEE Instrumentation and
  Measurement Technology Conference}.\hskip 1em plus 0.5em minus 0.4em\relax
  IEEE, 2008, pp. 985--990.

\bibitem{WirAdv}
M.~Iker, D.~Tobias, S.~Peter, B.~Josef, and P.~Alexander, ``An overview of
  technical challenges and advances of inductive wireless power transmission,''
  \emph{Proceedings of the IEEE}, vol. 101, no.~6, pp. 1302--1311, 2013.

\bibitem{AuCS}
B.~Walzel, C.~Sturm, J.~Fabian, and M.~Hirz, ``Automated robot-based charging
  system for electric vehicles,'' in \emph{16. Internationales Stuttgarter
  Symposium}.\hskip 1em plus 0.5em minus 0.4em\relax Springer, Wiesbaden, 2016,
  pp. 937--949.

\bibitem{AutAlign}
I.~Cortes and K.~Won-jong, ``Automated {A}lignment with {R}espect to a {M}oving
  {I}nductive {W}ireless {C}harger,'' in \emph{IEEE Transactions on
  Transportation Electrification}, 2021.

\bibitem{touchvr}
D.~Trinitatova and D.~Tsetserukou, ``Touch{VR}: a {W}earable {H}aptic
  {I}nterface for {VR} {A}imed at {D}elivering {M}ulti-modal {S}timuli at the
  {U}ser’s {P}alm,'' in \emph{SIGGRAPH Asia 2019 XR}, 2019, pp. 42--43.

\bibitem{yem2019}
V.~Yem, H.~Kajimoto, K.~Sato, and H.~Yoshihara, ``A {S}ystem of {T}actile
  {T}ransmission on the {F}ingertips with {E}lectrical-{T}hermal and
  {V}ibration {S}timulation,'' in \emph{International Conference on
  Human-Computer Interaction}.\hskip 1em plus 0.5em minus 0.4em\relax Springer,
  Cham, 2019, pp. 101--113.

\bibitem{xgboost}
\BIBentryALTinterwordspacing
T.~Chen and C.~Guestrin, ``X{GB}oost: {A} {S}calable {T}ree {B}oosting
  {S}ystem,'' in \emph{Proceedings of the 22nd ACM SIGKDD International
  Conference on Knowledge Discovery and Data Mining}, ser. KDD '16.\hskip 1em
  plus 0.5em minus 0.4em\relax New York, NY, USA: Association for Computing
  Machinery, 2016, p. 785–794. [Online]. Available:
  \url{https://doi.org/10.1145/2939672.2939785}
\BIBentrySTDinterwordspacing

\bibitem{batch}
S.~Ioffe and C.~Szegedy, ``Batch normalization: Accelerating deep network
  training by reducing internal covariate shift,'' in \emph{International
  conference on machine learning.}\hskip 1em plus 0.5em minus 0.4em\relax PMLR,
  2015, pp. 448--456.

\bibitem{batch_why}
S.~Santurkar, D.~Tsipras, A.~Ilyas, and A.~Madry, ``How does batch
  normalization help optimization?'' in \emph{Proceedings of the 32nd
  International Conference on Neural Information Processing Systems}, ser.
  NIPS'18.\hskip 1em plus 0.5em minus 0.4em\relax Red Hook, NY, USA: Curran
  Associates Inc., 2018, p. 2488–2498.

\bibitem{raschka2018model}
S.~Raschka, ``Model evaluation, model selection, and algorithm selection in
  machine learning,'' \emph{arXiv preprint arXiv:1811.12808}, 2018.

\end{thebibliography}

\end{document}